\documentclass[10pt,twocolumn,letterpaper]{article}

\usepackage{cvpr}              

\usepackage{graphicx}
\usepackage{amsmath}
\usepackage{amssymb}
\usepackage{booktabs}
\usepackage{bbding}

\usepackage[pagebackref,breaklinks,colorlinks]{hyperref}

\usepackage[capitalize]{cleveref}
\crefname{section}{Sec.}{Secs.}
\Crefname{section}{Section}{Sections}
\Crefname{table}{Table}{Tables}
\crefname{table}{Tab.}{Tabs.}

\def\cvprPaperID{6662} 
\def\confName{CVPR}
\def\confYear{2023}

\newcommand{\dataset}{\textit{\text{BracketFlare}}}

\begin{document}

\title{Nighttime Smartphone Reflective Flare Removal \\Using Optical Center Symmetry Prior}

\author{
Yuekun Dai$\,\,\,\,$ Yihang Luo $\,\,\,\,$ Shangchen Zhou $\,\,\, $Chongyi Li\thanks{Corresponding author.} $\,\,\, $ Chen Change Loy \\
S-Lab, Nanyang Technological University \\
\texttt{\small \{YDAI005, c200211, s200094, chongyi.li, ccloy\}@ntu.edu.sg}\\ \vspace{-6mm}
{\tt\small \url{https://ykdai.github.io/projects/BracketFlare}}
}

\maketitle

\begin{abstract}

Reflective flare is a phenomenon that occurs when light reflects inside lenses, causing bright spots or a ``ghosting effect'' in photos, which can impact their quality. Eliminating reflective flare is highly desirable but challenging. Many existing methods rely on manually designed features to detect these bright spots, but they often fail to identify reflective flares created by various types of light and may even mistakenly remove the light sources in scenarios with multiple light sources. To address these challenges, we propose an optical center symmetry prior, which suggests that the reflective flare and light source are always symmetrical around the lens's optical center. This prior helps to locate the reflective flare's proposal region more accurately and can be applied to most smartphone cameras.
Building on this prior, we create the first reflective flare removal dataset called BracketFlare, which contains diverse and realistic reflective flare patterns. We use continuous bracketing to capture the reflective flare pattern in the underexposed image and combine it with a normally exposed image to synthesize a pair of flare-corrupted and flare-free images. With the dataset, neural networks can be trained to remove the reflective flares effectively. Extensive experiments demonstrate the effectiveness of our method on both synthetic and real-world datasets.
\vspace{-4mm}

\end{abstract}

\section{Introduction}
\label{sec:intro}

\begin{figure}[t]
  \centering
   \includegraphics[width=1.0\linewidth]{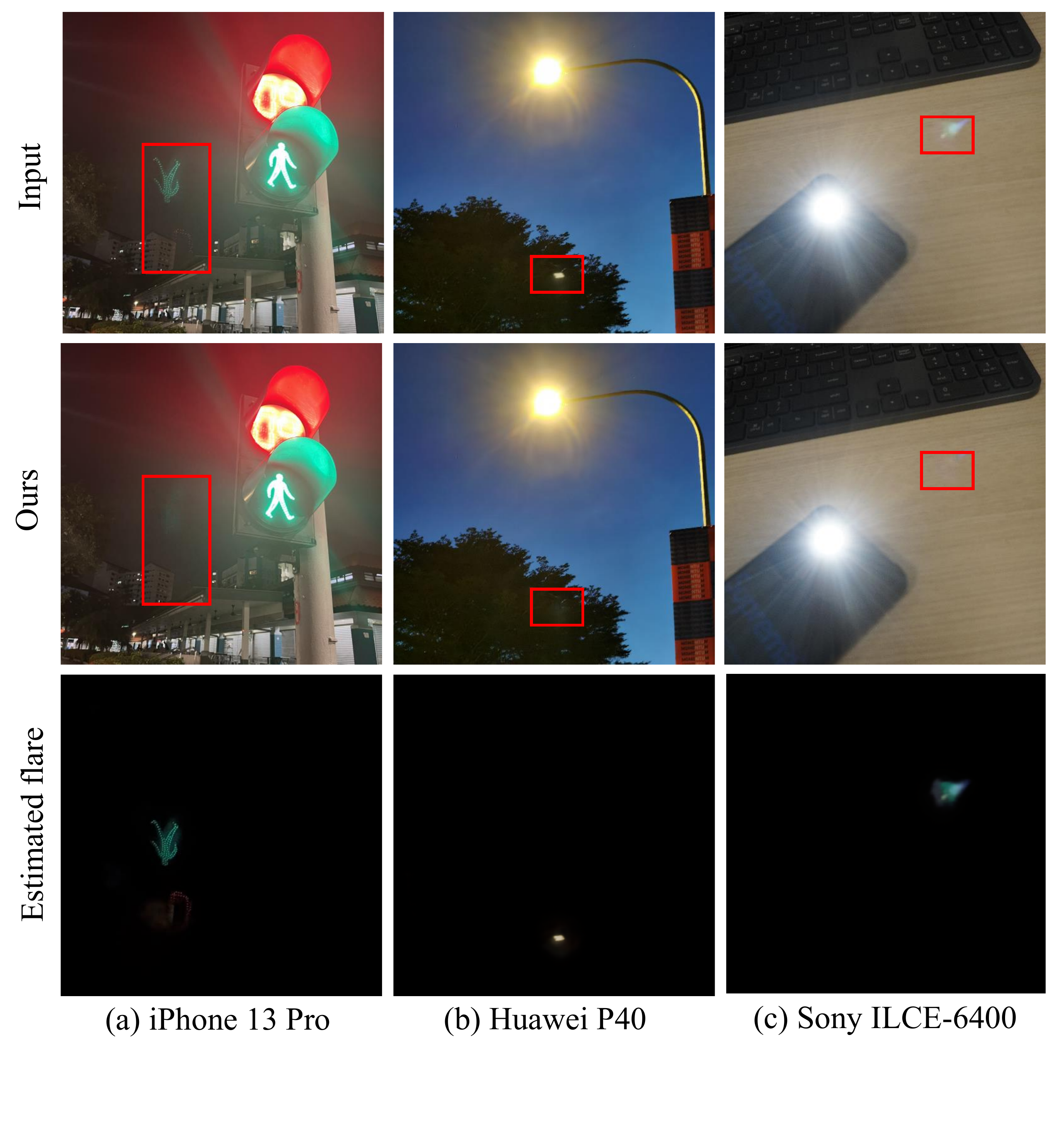}
   \vspace{-13mm}
   \caption{Reflective flare removal with our proposed dataset and prior. The network trained on our dataset can effectively separate the reflective flare and flare-free images from a flare-corrupted image. Our method can generalize well to different types of smartphones and even cameras with specific types of lenses.}
   \vspace{-6.5mm}
   \label{fig:reflective_flare_removal}
\end{figure}

\begin{figure}[t]
  \centering
   \includegraphics[width=1.0\linewidth]{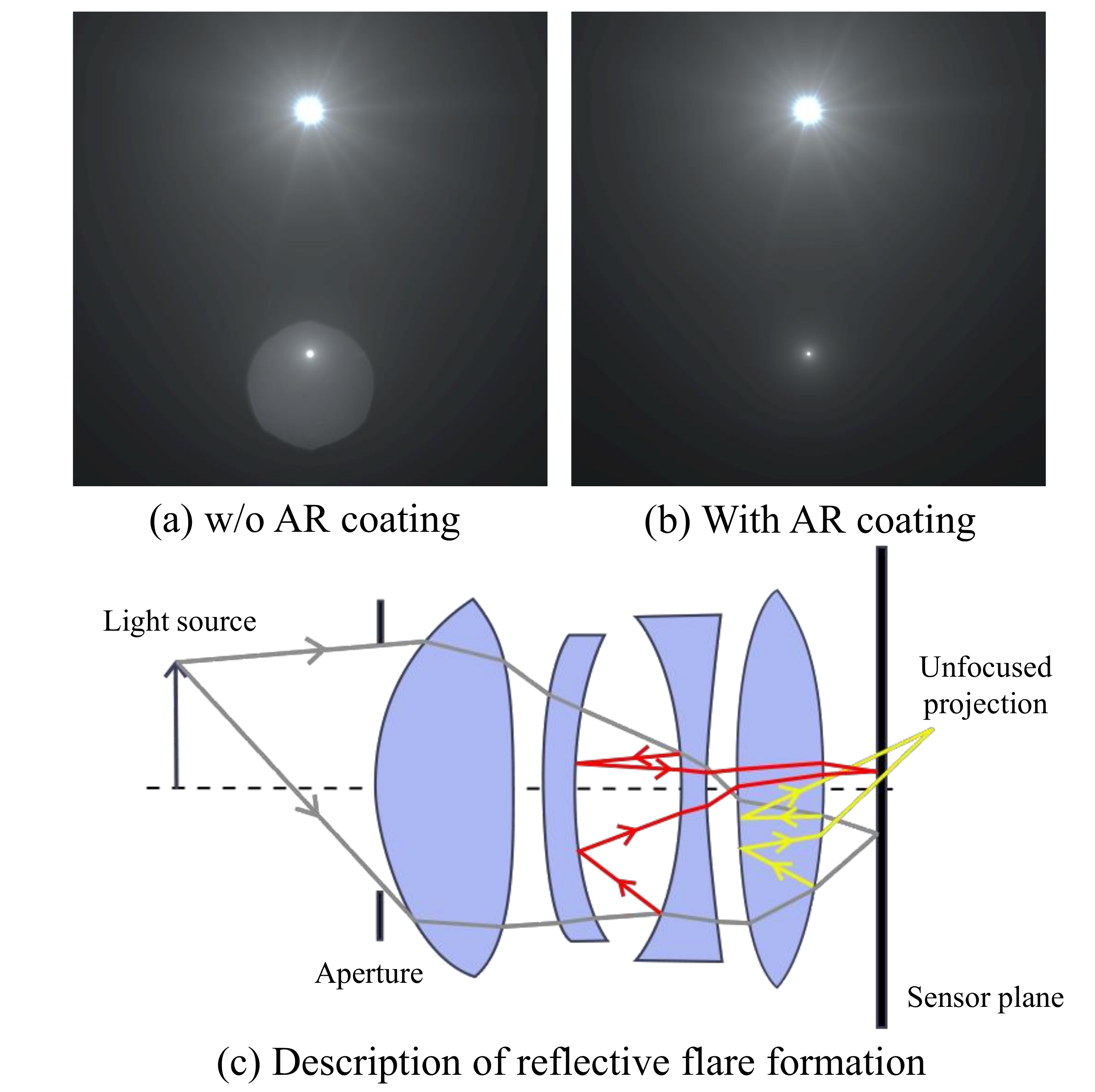}
   \vspace{-6mm}
   \caption{Schematic diagram of reflective flare formation. Reflective flares are generated by multiple reflections inside the lens.
    In (c), yellow lines show the generation principle of the
    aperture-shaped reflective flare. Red lines represent the formation of the bright spot. Since the unfocused reflective flare (yellow
    line, e.g., (a)) can be effectively alleviated by AR coating. In this work, we mainly focus on removing the focused reflective flare (red line, e.g., (b)) from night images.} 
   \vspace{-5mm}
   \label{fig:flare_generation}
\end{figure}

Lens flare artifacts can occur when a strong light enters the camera's field of view, causing scattering and reflection within the lenses. 
Unlike radial-shaped scattering flares, reflective flares appear as groups of bright spots~\cite{physically,dai2022flare7k} or a ``ghosting effect" like the examples in Fig.~\ref{fig:reflective_flare_removal}. 
With multiple reflections, the light source creates multiple projections at various positions behind the lens. 
If the projection is far from the focal plane, the defocused effect can brighten the whole sensor and degrade the photo's quality as shown in Fig.~\ref{fig:flare_generation}.
In near-focus situations, images of the light source can result in a line of unfocused, bright blobs on the captured photo.

Applying anti-reflective (AR) coating is a common technique for reducing internal reflections~\cite{wu2021train}. The coating is effective for unfocused spots with low luminance, but cannot mitigate a bright spot with high luminance at the focal plane. Even with a well-designed lens system, a bright reflective spot may remain in the photo, known as the ghosting effect~\cite{practical}. 
Reflective flare is common on all types of smartphone lenses, and certain types of smartphones (e.g., the iPhone series) can produce a large area of inverted reflective flare from neon lights or large, luminous standing signs, as shown in Fig.~\ref{fig:reflective_flare_removal}. 
The phenomenon becomes more pronounced when capturing a photo at night with multiple lights in the scene.  
Numerous bright spots and inverted reflections manifest and propagate in different directions with the lens orientation. Therefore, reflective flare removal algorithms are highly desirable.

Removing reflective flare is a difficult task due to the vast diversity of reflective flares that can be caused by different types of light sources and lenses. 
Reflective flares often appear as bright spots that can resemble streetlights from a distance, making it challenging to differentiate between reflective flares and other light sources. 
Current methods for removing flare spots mainly rely on feature detection techniques like Speeded-Up Robust Features~\cite{surf} (SURF) or Scale-Invariant Feature Transform~\cite{sift} (SIFT) to extract spot regions~\cite{auto_removal,auto_removal2,automated_removal}.
These methods, however, struggle to effectively detect blobs on complex backgrounds and only work for specific spot patterns. Moreover, these methods are ineffective in removing reflective flares caused by lamp tubes and LED matrices.
Many learning-based methods and datasets for reflective flare removal have been proposed~\cite{wu2021train,dai2022flare7k,light_source,rank_1}. Although these datasets contain subsets of reflective flares, they are not sufficient for good generalization.
For instance, the reflective flare subset from Wu \etal~\cite{wu2021train} is captured using the same lens and light source, making it difficult to generalize to different flares in real-world scenarios. 
Dai \etal~\cite{dai2022flare7k} propose a synthetic dataset with 10 types of reflective flares, but training on this dataset tends to mislead a network to erroneously remove other tiny bright spots like street lights and bright windows.
As a result, it remains challenging for existing algorithms to accurately detect and remove various types of reflective flares while preserving details in other regions.

To overcome the challenges of detecting and removing reflective flares, we propose a novel optical center symmetry prior, which suggests that the main spot of the reflective flare and the light source are always symmetrical around the lens's optical center in captured images. 
This prior applies to most smartphone cameras and some professional cameras. With this prior, we can easily detect the possible locations of reflective flares.
We also found that the flare spots always have the same patterns as the brightest regions of the light source. Therefore, we can synthesize reflective flares by using the light source's patterns at low exposure. Based on these characteristics of reflective flares, we develop the first reflective flare removal dataset, named \dataset.
Specifically, we use continuous bracketing to capture the light source's pattern on an underexposed image and select the normally exposed image to create the light source and the natural image. 
Next, we rotate the light source pattern 180 degrees around the optical center, apply blur and color augmentation, and use it to create synthetic reflective flares. 
We can add the reflective flare to the normally exposed image to produce paired flare-corrupted and flare-free images.

Thanks to the optical center symmetry prior, we further propose an end-to-end pipeline for reflective flare removal, which can easily apply standard neural networks designed for image restoration. 
Specifically, to demonstrate the effectiveness of our dataset and pipeline, we train two CNN-based networks~\cite{HINet,MPRNet} and two Transformer-based networks~\cite{Restormer,Uformer} as baselines. These models trained on our dataset effectively suppress the reflective flare and generalize well to real-world scenarios.
%
The contributions of our work are as follows:
\begin{itemize}
    \item We propose a new optical center symmetry prior that uncovers crucial properties of reflective flares on smartphones' cameras, including their locations of occurrence and patterns.
    \item Based on this prior, we explore a new method for constructing a reflective flare removal dataset called \dataset. This dataset includes various types of light sources and diverse indoor and outdoor scenes, providing a solid foundation for this task. 
    \item We design an end-to-end reflective flare removal pipeline that incorporates the optical center symmetry prior to both training data generation and network structures. Extensive experiments demonstrate the essential role of our dataset and prior in reflective flare removal, which effectively generalizes to different night scenes
\end{itemize}

\section{Related Work}
\label{sec:formatting}

\noindent{\bf Scattering flare removal.}
Most flare removal methods mainly focus on removing the scattering flares and glare effects.
Based on the glow smooth prior, unsupervised methods \cite{jin2022unsupervised,nighttime_sharma} are proposed to separate the glare effect from the flare-corrupted images.
For supervised flare removal, Wu \etal~\cite{wu2021train} and Dai \etal~\cite{dai2022flare7k} have respectively proposed two flare image datasets.
These flare images can be added to flare-free images to synthesize the flare-corrupted and flare-free image pairs, making training an end-to-end flare removal network possible.
Due to the difficulty in collecting real paired data, Qiao \etal~\cite{light_source} designed a pipeline to train the flare removal network with unpaired data following the idea of Cycle-GAN~\cite{CycleGAN}.
Besides, artifacts from under-display cameras~\cite{feng2021removing} and nighttime haze~\cite{li_nighttime_2015} share similar attributes with the scattering flares. Thus, the related methods can also alleviate scattering flares artifacts.

\vspace{2mm}
\noindent{\bf Reflective flare removal.}
Traditional reflective flare removal is typically accomplished through feature detection and image inpainting.
Flare detection in these approaches~\cite{auto_removal,auto_removal2,automated_removal,aerial_tracking} assumes that reflective flares are bright spots with specific shapes.
Such an assumption is fragile given nighttime situations with diverse light sources.
Since reflective flare removal can be viewed as a specific kind of reflection removal~\cite{zhang2018single}, one can adopt learning-based methods. 
However, existing datasets for this task, such as those proposed by Wu \etal~\cite{wu2021train}, Dai \etal~\cite{dai2022flare7k}, and Qiao \etal~\cite{light_source}, are not specifically designed for reflective flare removal and lack diversity in terms of light sources. 
These limitations hinder deep learning-based methods from effectively addressing real-world reflective flares, and may even lead to the removal of other tiny bright spots such as street lights.


\section{Optical Center Symmetry Prior}

\subsection{Definition for the prior}

\begin{figure}[t]
  \centering
   \includegraphics[width=0.9\linewidth]{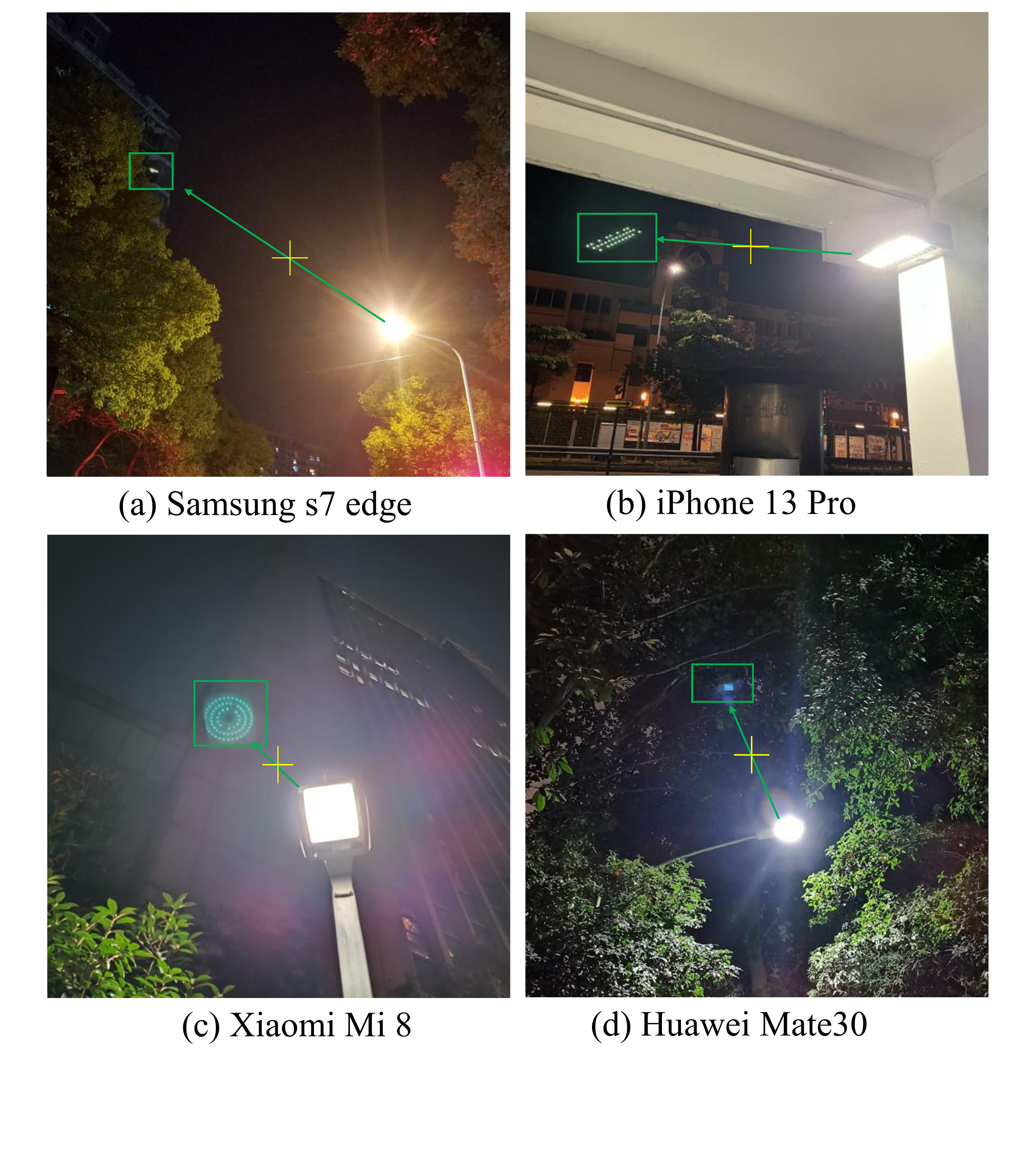}
   \vspace{-8mm}
   \caption{Strong light source may project a bright reflective flare on the photo. The reflective flares and corresponding light sources are symmetric around the optical center for most smartphone cameras. The yellow cross represents the optical center. 
   }
   \vspace{-4mm}
   \label{fig:def_reflective_flare}
\end{figure}

When pointing the camera at a light source, strong light can sometimes create lens flares in photos. 
In well-designed lens systems with AR coatings, reflective flares appear as bright spots on the photo, rather than generating a light fog. 
Through our investigation of most smartphones' main cameras, we discover that the bright spots of reflective flares and light sources are always symmetrical around the lens's optical center, as shown in Fig.~\ref{fig:def_reflective_flare}. 
By aligning the light source and optical center, we can easily locate the reflective flares, eliminating the need to search the entire image for possible occurrences.
Besides, since the main bright spots are caused by focused reflective flares, they share the same attributes as the light source. For example, the reflective flares may display clear details for each small LED lamp bead, as shown in Fig.~\ref{fig:def_reflective_flare}(b) and Fig.~\ref{fig:def_reflective_flare}(c).

\subsection{Physical formulation}

\begin{figure}[t]
  \centering
   \includegraphics[width=1.0\linewidth]{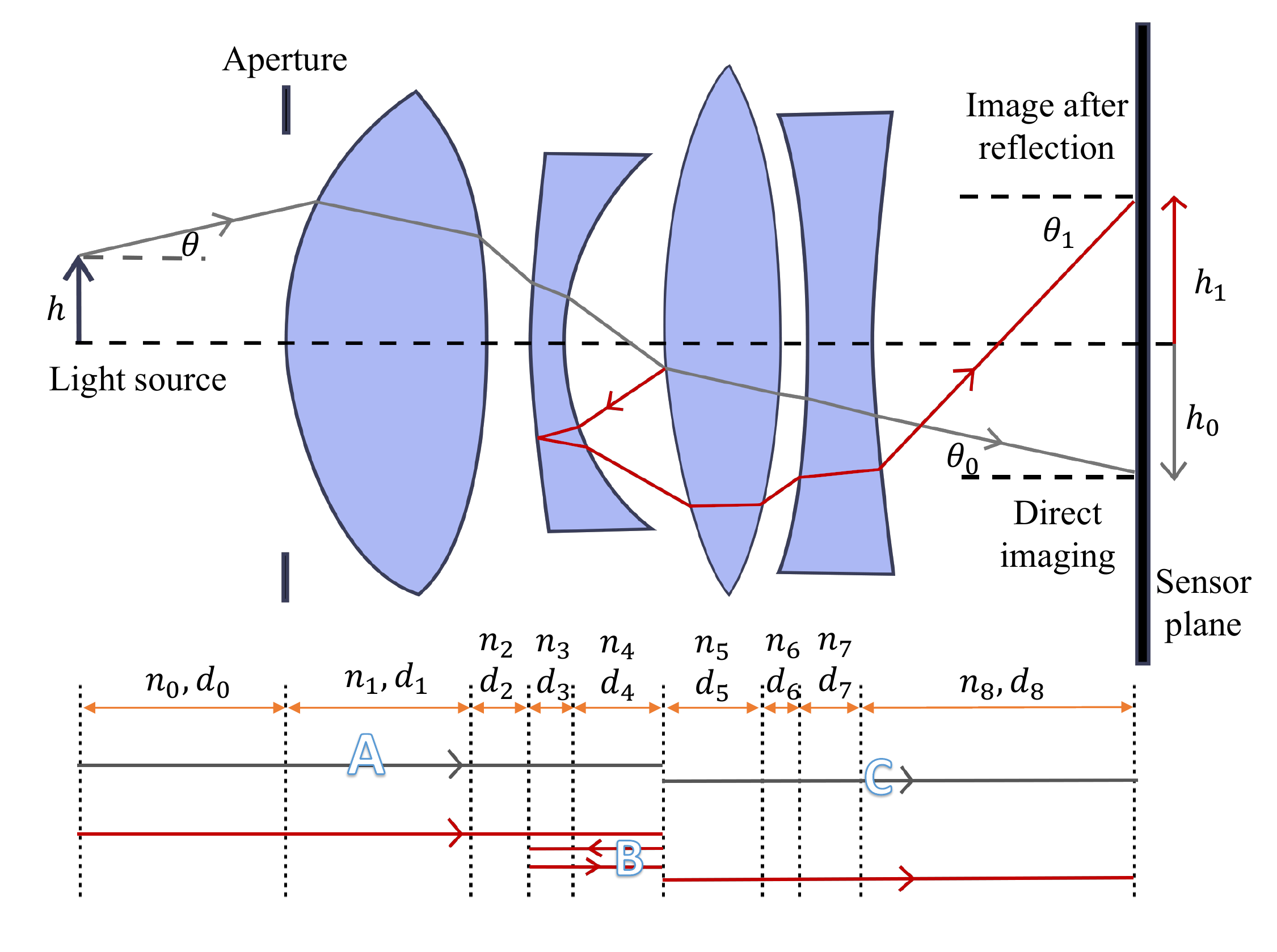}
   \vspace{-8mm}
   \caption{Physical explanation of reflective flare formation. In this figure, $h$ and $\theta$ indicate the height and the angle of incidence of the light source.  $d$ and $n$ denote the thickness and refractive index of each layer. $A$, $B$, and $C$ represent the transfer matrices for each part of the lens system. The red line represents the imaging after reflection and the grey line shows the path of direct imaging. This figure depicts the situation with reflection in the 3rd and 5th air-lens layer, in real-world situations, the reflections can happen at any place including the lens protector and CMOS layer.}
   \vspace{-4mm}
   \label{fig:flare_physics}
\end{figure}

To illustrate the formation of optical center symmetry prior, we suppose that an optical ray $\mathbf{r}$ is emitted from the light source, as shown in Fig. \ref{fig:flare_physics}. This ray can be described as $\mathbf{r}=[h \ \theta]^\top$. The states of this ray after each lens surface can be calculated by multiplying transfer matrices~\cite{pedrotti2017introduction,practical}. We define the translation matrix at layer $i$ as $\mathbf{T_i}$, and the refraction matrix at spherical surface with radius $R_i$ and refractive index $n_i,n_{i+1}$ as $\mathbf{F_i}$. For the reflection at the $i$th layer with radius $R_i$, the reflection matrix can be stated as $\mathbf{R_i}$. These matrices can be stated as:
\begin{scriptsize}
\begin{equation}
\label{eq1}
    \mathbf{T_i}=
    \begin{pmatrix} 
    1 & d_i \\ 
    0 & 1 
    \end{pmatrix},
    \mathbf{R_i}=
    \begin{pmatrix} 
    1 & 0 
    \\ \frac{2}{R_i} & 0 
    \end{pmatrix},
    \mathbf{F_i}=\begin{pmatrix} 
    1 & 0 \\ 
    \frac{n_{i+1}-n_i}{n_{i+1}R_i} & \frac{n_i}{n_{i+1}} 
    \end{pmatrix}
\end{equation}
\end{scriptsize}

In Fig.~\ref{fig:flare_physics}, $h_1$ represents the position of a real image that focuses on the sensor through the lens' internal reflection. We define the ray's forward propagation matrix as $\mathbf{A}$ and $\mathbf{C}$ and the reflection process as $\mathbf{B}$. They are defined as:
\begin{equation}
\label{eq2}
\begin{split}
\begin{aligned}
    A=&\mathbf{T_4}\mathbf{F_3}\mathbf{T_3}\mathbf{F_2}\mathbf{T_2}\mathbf{F_1}\mathbf{T_1}\mathbf{F_0}\mathbf{T_0}\\
    B=&\mathbf{T_4}\mathbf{F_3}\mathbf{T_3}\mathbf{R_2}^{-1}\mathbf{T_3}\mathbf{F_3}^{-1}\mathbf{T_4}\mathbf{R_4}\\
    C=&\mathbf{T_8}\mathbf{F_8}\mathbf{T_7}\mathbf{F_7}\mathbf{T_6}\mathbf{F_5}\mathbf{T_5}\mathbf{F_4}\\
\end{aligned}
\end{split}
\end{equation}

With these matrices, the ray after reflection in the focal plane $\mathbf{r_1}$ and direct imaging $\mathbf{r_0}$ can be written as:
\begin{equation}
\label{eq3}
\begin{split}
\begin{aligned}
    \begin{pmatrix} 
    h_1 \\ \theta_1
    \end{pmatrix}=
    \mathbf{CBA}\begin{pmatrix} 
    h \\ \theta
    \end{pmatrix},
    \begin{pmatrix} 
    h_0 \\ \theta_0
    \end{pmatrix}=
    \mathbf{CA}\begin{pmatrix} 
    h \\ \theta
    \end{pmatrix}
\end{aligned}
\end{split}
\end{equation}

For the focused image, $h_0$ and $h_1$ will not vary with the change of the angle of incidence $\theta$.
Thus, we can deduce from Eq.~\eqref{eq3} that both $h_0$ and $h_1$ are linear with $h$.
This means that the line between the light source and the reflective flare will pass through the optical center and the ratio between the $h_0$ and $h_1$ is a constant.
We define this ratio as $k$ with $h_1=k h_0$. 
Thus, we can calculate the position of the reflective flare, if this constant $k$ is known.
Due to the lens design for most smartphone cameras, the value of $k$ is always -1.
It leads to the optical center symmetry prior stated above.

\begin{figure}[t]
  \centering
   \includegraphics[width=0.85\linewidth]{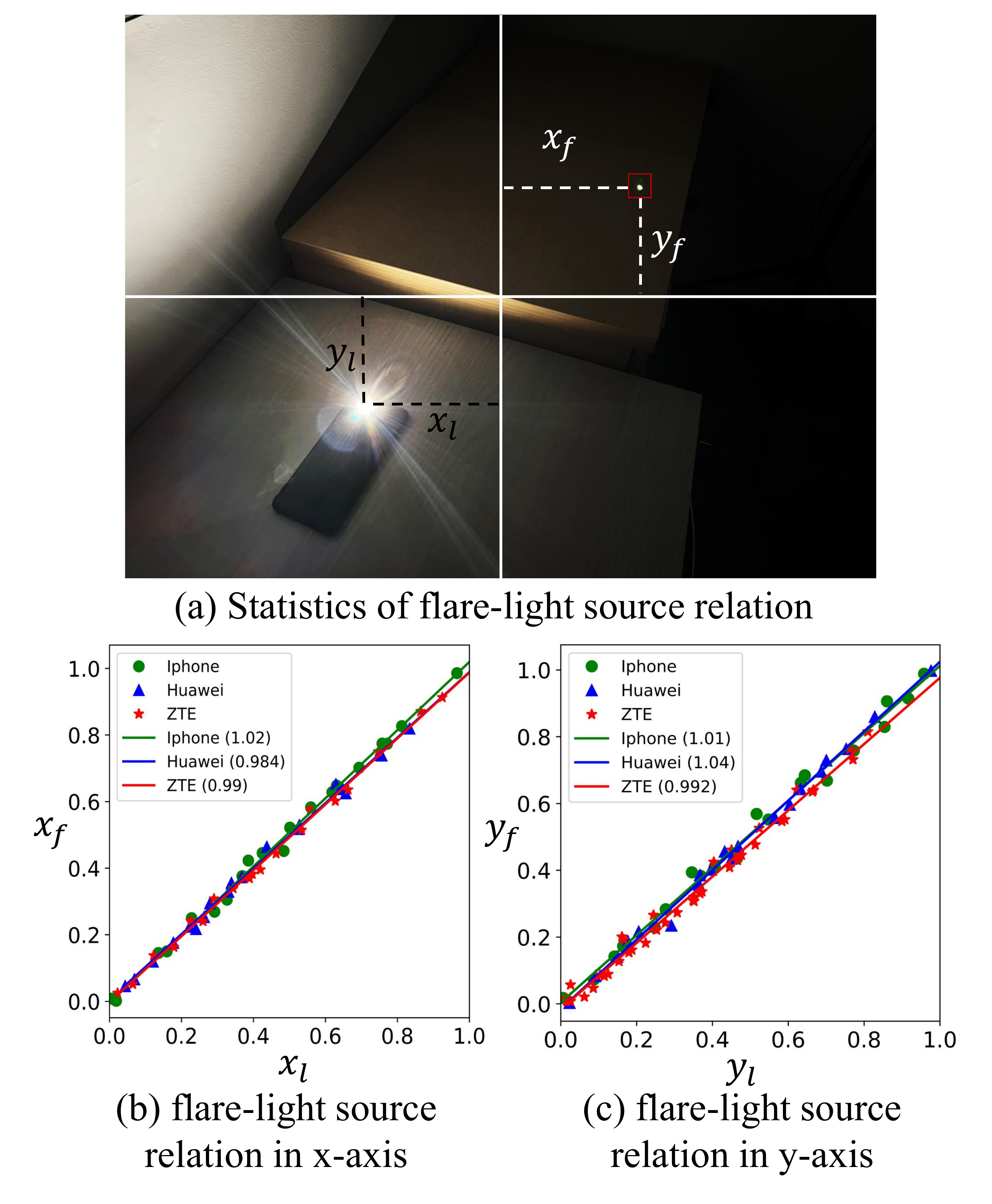}
   \vspace{-4mm}
   \caption{ Experimental statistics of the optical center symmetry prior. We capture images with reflective flares in a dark environment and mark the position of the light source and reflective flares. $x_f$ and $x_l$ represent the x-coordinate of flare and light source and so as $y_f$ and $y_l$. The origin is set to be the center of the image. The relationship between the flare and the light source's absolute values of coordinates are plotted above. From figures (b) and (c), the coordinate relationships are linear and the slope is 1, which confirms our prior.}
   \vspace{-8mm}
   \label{fig:statistics}
\end{figure}

\begin{figure*}[t]
   \includegraphics[width=0.95\linewidth]{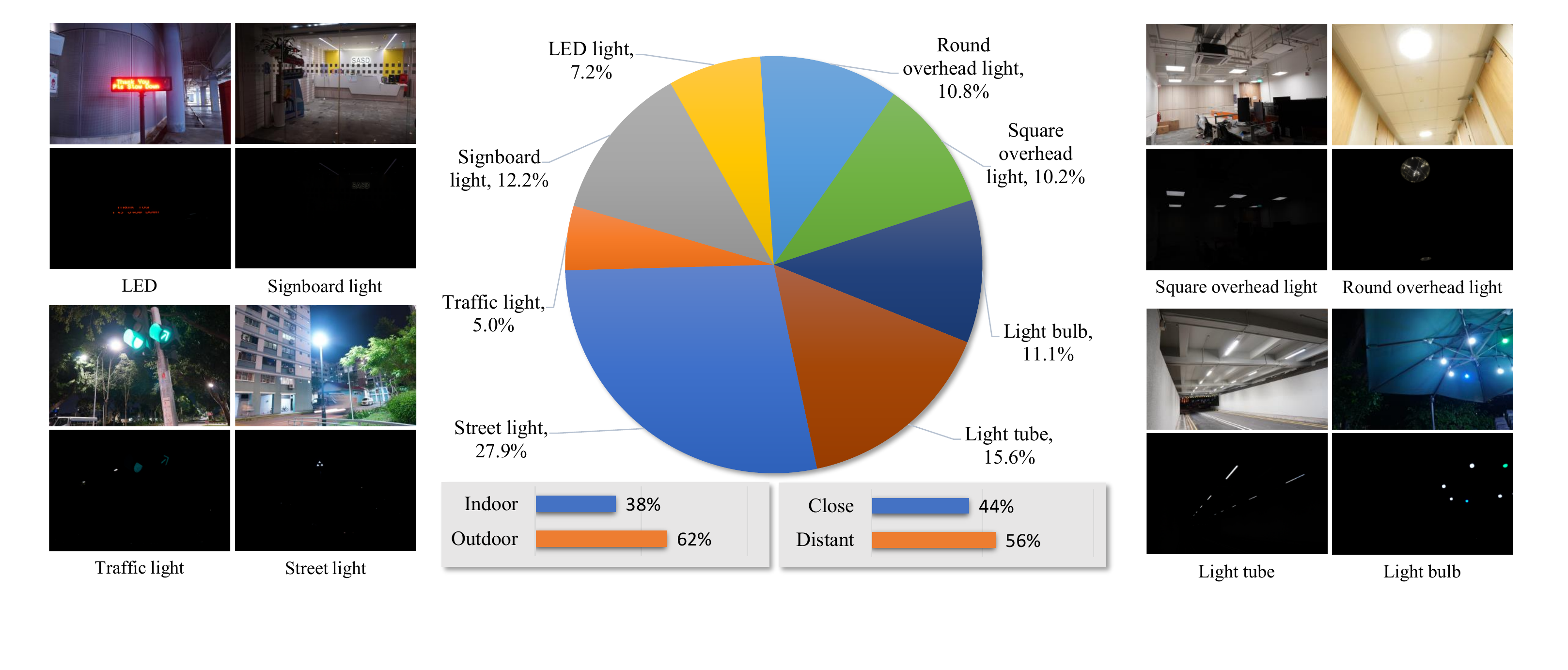}
   \centering
   \vspace{-10mm}
  \caption{Visualization and distribution of our dataset. It contains different types of light sources in diverse scenes. Based on light source types, we classify them into eight categories with different flare patterns.}
    \vspace{-2mm}
  \label{fig:data_review}
\end{figure*}

\subsection{Experimental statistics}

To demonstrate the effectiveness of our proposed prior in real-world situations, we conducted experiments using the main cameras of three smartphones: the Huawei P40, iPhone 13 Pro, and ZTE Axon 20 5G. In dark scenes, we captured 20 photos using each camera and marked the positions of the light source and reflective flare relative to the center of the image. As shown in Fig.~\ref{fig:statistics}, we observed that for all three cameras, the position of the reflective flare is linearly related to the position of the light source, with a slope of 1.
However, since the light source may often create overexposed regions, manually marking the position of the light source can still introduce some errors. Considering this effect, we conclude that the results from all three phones align with our proposed prior.

\section{Data Collection}
\label{data_collection}

Although the optical center prior can guide us in finding the reflective flare, the lack of paired images still presents a challenge for training an effective model for flare removal. As reflective flares are caused by the lens system, it is physically impossible to capture real flare-free and flare-corrupted image pairs. Moreover, given the huge variety of nighttime lights, it would be tedious to synthesize reflective flares for each type of light source.

To address the challenge of generating paired images for reflective flare removal, we propose a semi-synthetic dataset, named \dataset. 
Since reflective flares always have the same pattern as the brightest region of the light source, we reduce the exposure time to capture the patterns of the light source's brightest region, which are used to synthesize the reflective flares. 
Specifically, we mount a professional camera (Sony A7 III) on a tripod and use continuous bracketing to capture five photos of the same static scene with a step of 3 EV as a group. 
We then select the normally exposed image and the lowest exposure image from each group to compose the pair. 
To increase the diversity and realism of our dataset, we choose scenes with different light sources that are bright enough to leave reflective flares.
Finally, we collect a dataset consisting of 440 pairs of 4K images as training data and 40 images as test data. A general overview of our dataset is depicted in Fig.~\ref{fig:data_review}, showing that the dataset contains diverse and commonly seen flare patterns.

\section{Proposed Method}
\subsection{Postprocessing our dataset}
\label{sec:data_augment}

\begin{figure*}[t]
   \centering
   \includegraphics[width=0.85\linewidth]{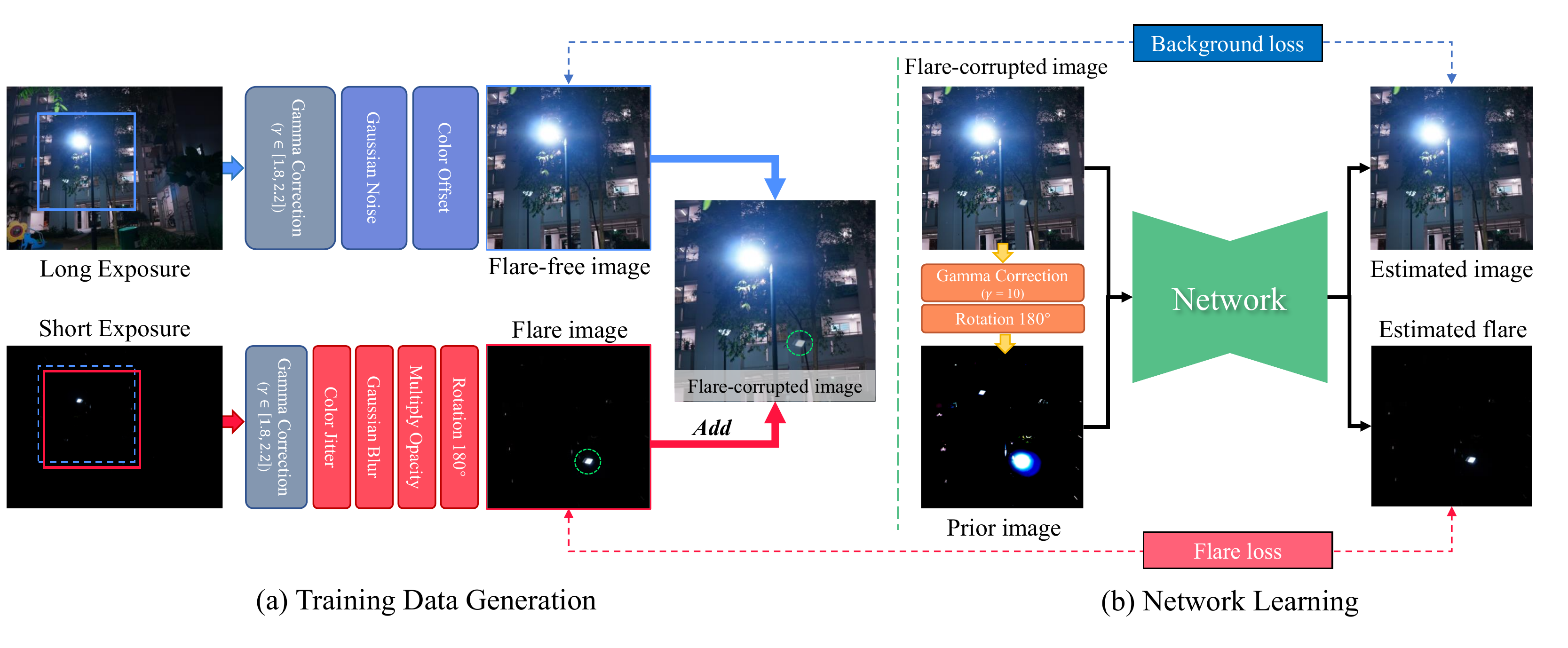}
   \vspace{-5mm}
  \caption{The pipeline of our data processing module and flare removal network. The flare image and a flare-free image from our paired dataset are resized and randomly cropped. After data augmentation, they are added to synthesize the corresponding flare-corrupted image. To encode the optical symmetry prior, the input flare-corrupted image is center flipped as the prior image. We also apply gamma correction to extract the light source region. Then, the prior image is concatenated with the flare-corrupted image as the input with 6 channels. Finally, the network predicts an output with 6 channels (a flare-free image with 3 channels and an estimated flare with 3 channels).}
  \vspace{-2mm}
  \label{fig:pipeline}
\end{figure*}

To increase the diversity of our dataset, we generate paired flare-corrupted and flare-free images on the fly with augmentation. 
As depicted in Fig.~\ref{fig:pipeline}(a), we first resize the 4K image to 800$\times$1200. 
To simulate the possible misalignment of the image's center and optical center, we introduce a slight translation with $t\sim U(0,15)$ pixels to the underexposed light source image. 
Then, we randomly crop the paired light source image and the normally exposed image with a size of 512$\times$512. 
Since our captured images are already gamma-encoded, we follow the settings of Wu~\textit{et~al.} \cite{wu2021train} and linearize them by applying a gamma correction, i.e., $I_g = I^\gamma$ with $\gamma$ randomly sampled from $[1.8,2.2]$. All our operations are performed in the linear space.

\noindent{\bf Synthetic reflective flare.} 
As lenses reflect light of different wavelengths with varying reflectivity, reflective flares often appear in different colors. To replicate this, we use a random color factor to multiply the low-exposed image. To simulate an unfocused situation, we also add random blur. In accordance with the optical center symmetry prior, we rotate the light source image to synthesize the reflective flare. Finally, we obtain the synthetic reflective flare by multiplying the flare image with an opacity.

\noindent{\bf Flare-free image.} 
The ISO is fixed for all images in our continuous bracketing. 
Therefore, our normally exposed images always share a long exposure time and have little noise. 
To simulate real night scenes, we add Gaussian noise to the images.
Finally, we randomly apply a small color offset to the flare-free image.
These operations aid in simulating light fog and glare effects in real-world scenarios.
\noindent{\bf Flare-corrupted image.} 
To create flare-corrupted images, we add the synthetic reflective flare to the flare-free images. Then, we apply gamma correction to the images with the gamma value obtained by taking the reciprocal of the previously sampled gamma correction value $\gamma$.
The resulting gamma-encoded flare-corrupted image, flare-free image, and flare image are used to train the neural networks.

\subsection{Flare removal pipeline}
As the optical center symmetry prior is a global prior, it is difficult for CNN-based networks to learn it, and even Transformer-based networks struggle to learn such symmetry. Therefore, it is crucial to incorporate this prior into the network with specialized designs.

To improve the network's ability to learn the optical center symmetry prior, as shown in Fig.~\ref{fig:pipeline}(b), we apply a gamma correction of $I_g = I^\gamma$ to the input flare-corrupted image with $\gamma = 10$. This operation extracts the nearly saturated areas of the light source in each channel. We then rotate this nearly saturated image by 180 degrees to obtain a prior image, which is concatenated with the flare-corrupted image to form a six-channel input. This rotated prior image provides an initial approximation of the reflective flare's pattern, which we expect the network to refine and enhance. The network's output is set to six channels, with three channels for the estimated flare-free image and three channels for the estimated flare. Our synthetic data provides supervision for all six channels, which facilitates network training.

\subsection{Loss function}
With the supervision of flare image $I_F$ and flare-free image $I_B$, our final loss is the combination of background loss $\mathcal{L}_{bg}$, flare loss  $\mathcal{L}_{flare}$, and reconstruction loss $\mathcal{L}_{rec}$. It can be written as:
\begin{equation}
\label{eq5}
\mathcal{L}=\mathcal{L}_{rec}+\mathcal{L}_{flare}+\mathcal{L}_{bg}.
\end{equation}

We define our flare removal network as $f_\theta$. It takes flare-corrupted image $I_{input}$ and its prior image $I_{prior}$ as inputs. The flare-free image and flare can be estimated by:
\begin{equation}
\label{eq6}
\hat{I}_{B},\hat{I}_{F}=f_\theta(I_{input},I_{prior}).
\end{equation}

With the estimated flare and flare-free images, the $L_1$ reconstruction loss can be expressed as:
\begin{equation}
\label{eq7}
\mathcal{L}_{rec}=|I_{input}-\hat{I}_{F} \oplus \hat{I}_{B}|,
\end{equation}
where $\oplus$ denotes an element-wise addition operation in the linearized gamma-decoded domain.

For the background loss, it consists of a default $\mathcal{L}_1$, a region-aware masked $\mathcal{L}_{mask}$ term, and a perceptual term $\mathcal{L}_{vgg}$ with a pre-trained VGG-19 network \cite{vgg}.
Compared with previous works \cite{zhang2018single,wu2021train,dai2022flare7k}, the main difference is the $\mathcal{L}_{mask}$. 
Since the regions of reflective flares are always small, a simple $L_1$ loss may ignore these regions, leading to the local optimum.
To encourage the network to focus on restoring the flare-corrupted regions, we calculate the flare's mask $I_M$ ($>0.2$) from the ground truth of flare image $I_F$. 
This masked loss can be written as:
\begin{equation}
\label{eq8}
\mathcal{L}_{mask}=|I_M*(\hat{I}_B-I_B)|.
\end{equation}

Overall, the  background loss $\mathcal{L}_{bg}$ is defined as:
\begin{equation}
\label{eq9}
\mathcal{L}_{bg}=w_1\mathcal{L}_1+w_2\mathcal{L}_{vgg}+w_3\mathcal{L}_{mask},
\end{equation}
where $w_1,w_2,w_3$ are respectively set to 0.5, 0.1, and 20.0  in our experiments.
Flare loss $\mathcal{L}_{flare}$ has the identical form as the background loss.

\begin{figure*}
   \centering
   \includegraphics[width=0.85\linewidth]{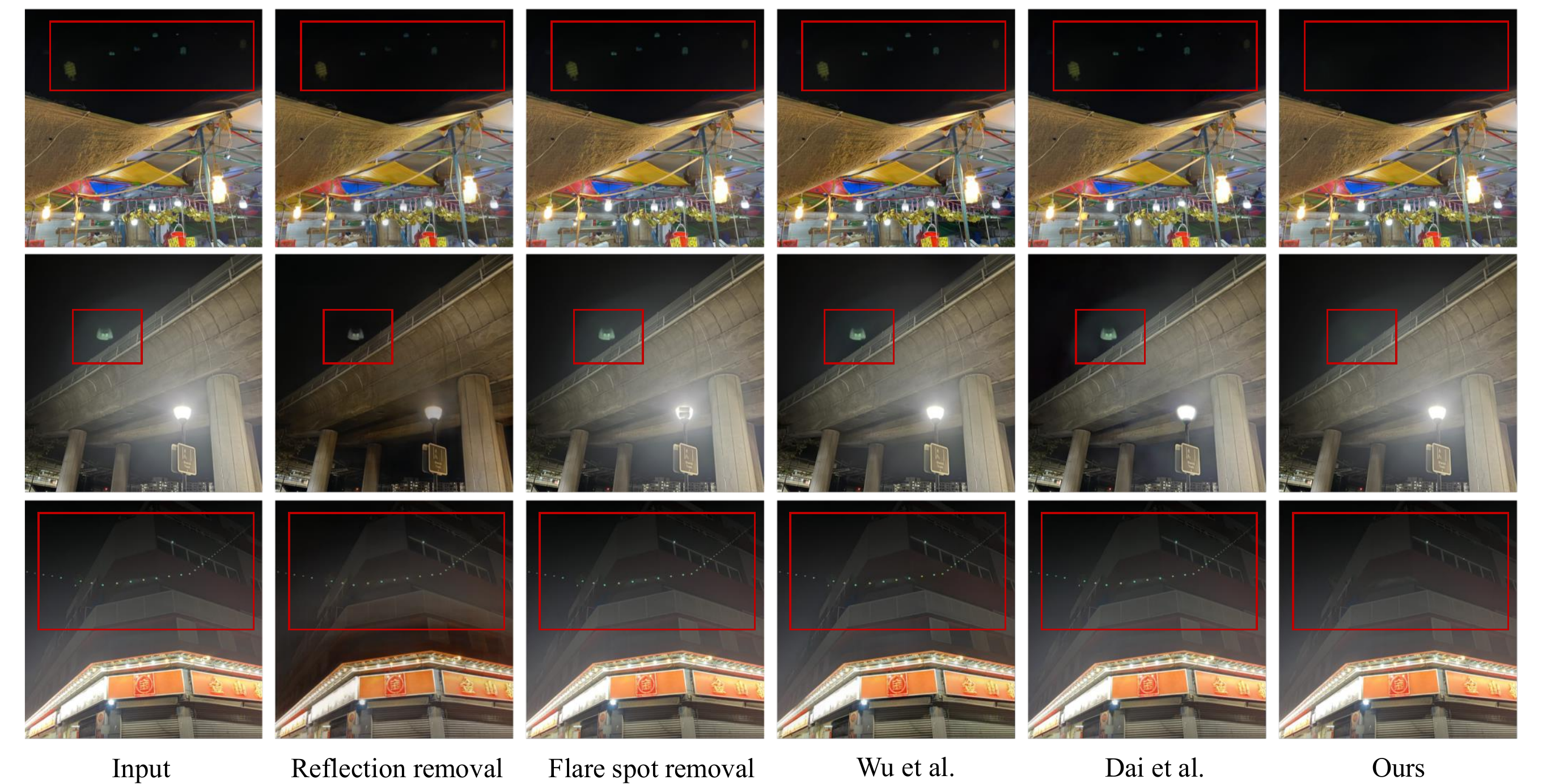}
   \vspace{-2mm}
  \caption{Visual comparison of different flare removal and reflection removal methods on real-world nighttime flare images.
  }
  \vspace{-3mm}
  \label{fig:previous_methods}
\end{figure*}

\begin{figure*}
   \centering
   \includegraphics[width=0.98\linewidth]{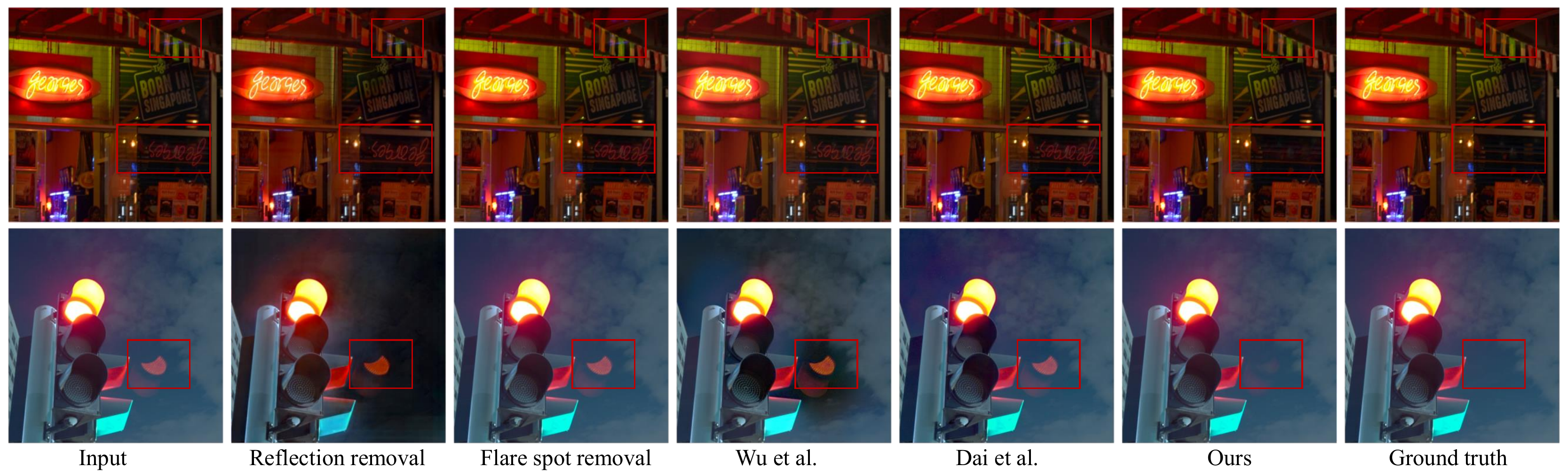}
   \vspace{-2mm}
  \caption{Visual comparison of different flare removal and reflection removal methods on synthetic nighttime flare images.
  }
  \vspace{-3mm}
  \label{fig:previous_methods_synthetic}
\end{figure*}

\section{Experiments}
\begin{figure*}
   \includegraphics[width=1.0\linewidth]{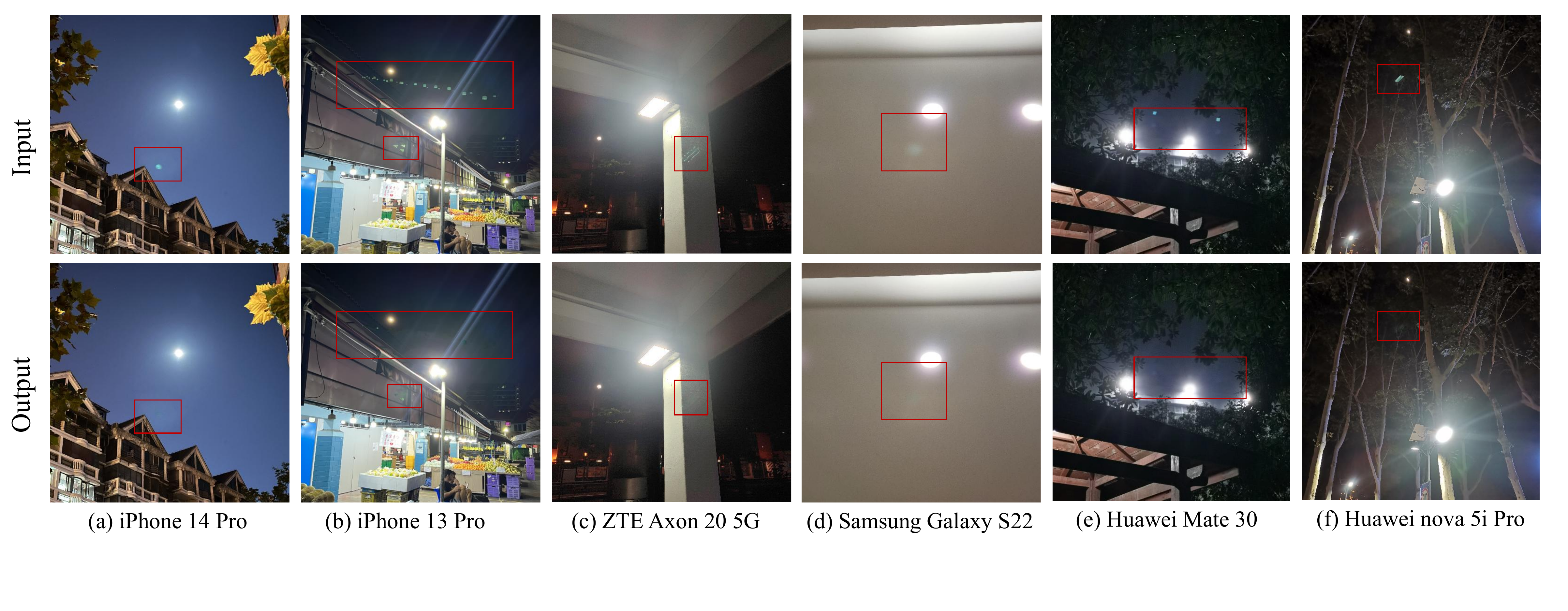}
   \vspace{-12mm}
  \caption{Visual comparison of flare removal on real-world nighttime flare images that are captured by different smartphones. }
  \label{fig:different_phones}
  \vspace{-4mm}
\end{figure*}

\noindent{\bf Evaluation on synthetic data.} To evaluate our method, we collect 40 paired data using the process mentioned in Sec.~\ref{data_collection} to generate a synthetic test dataset, in which ground truth images are available. We adopt PSNR, SSIM~\cite{ssim}, and LPIPS~\cite{lpips} as our evaluation metrics.

\noindent{\bf Evaluation on real data.} To show our method's generalization ability, we also collect a dataset with the photos captured by different types of mobile phones, including iPhone 12 pro, iPhone 13 pro, iPhone 14 pro, Huawei mate 30, Huawei nova 5i Pro, Huawei P40, Samsung Galaxy s22, Samsung Galaxy Note9, ZTE Axon 20 5G and Oppo Reno4 SE 5G.
For real cases, we cannot obtain the ground truth. 
Since reflective flares always locate in small regions, they are hard to be evaluated by existing non-reference image quality assessment.
These metrics emphasize more on global image quality and are not specially designed for our task.
Therefore, we provide more visual results and conduct a user study to evaluate the performance.

\begin{table}
  \centering
  \caption{Percentage of users favoring our results (our dataset 
  + MPRNet$^\dag$) vs. previous work. Our method outperforms existing methods on both indoor and outdoor images.}
  \vspace{-2mm}
  \resizebox{0.8\linewidth}{!} {
  \begin{tabular}{@{}lcccc@{}}
    \toprule
    Method & Outdoor & Indoor\\
    \midrule
    Reflection removal \cite{zhang2018single} &3.0\% &8.5\% \\
    Flare spot removal \cite{auto_removal}   &4.5\% &0.0\% \\
    Flare removal (Wu et al.)\cite{wu2021train} &0.5\% &3.5\% \\
    Flare removal (Dai et al.)\cite{dai2022flare7k} &1.5\% &2.5\% \\
    Ours &\textbf{90.5\%} &\textbf{85.5\%} \\
    \bottomrule
  \end{tabular}
  }
  \vspace{-6mm}
  \label{tab:user_study}
\end{table}

\subsection{Comparison with different networks}
To validate the performance of our pipeline, we train different CNN-based and Transformer-based networks with our data. 
We use the default network settings of MPRNet~\cite{MPRNet}, HINet~\cite{HINet}, and Uformer~\cite{Uformer} and reduce the feature channels of Restormer~\cite{Restormer} from  48 to 24 due to GPU memory constraint.
In our experiments, these models are trained on an NVIDIA Geforce RTX3090 with a batch size of 2.
We use Adam optimizer~\cite{adam} with a learning rate of 1e-4 and train a model for 500k iterations.
Besides, we adopt a MultiStepLR with a milestone of 200k and set the gamma to 0.5.
These models are evaluated using the synthetic test dataset as shown in Table \ref{tab:network}.
From experiments, we find that MRPNet$^\dag$~\cite{MPRNet} performs better than other networks.
Thus, we treat MPRNet$^\dag$~\cite{MPRNet} as our baseline method and conduct the following ablation study.

\vspace{-1mm}
\subsection{Comparison with previous works}
\vspace{-1mm}
We compare our method with different flare removal datasets~\cite{wu2021train,dai2022flare7k} and reflection removal dataset~\cite{zhang2018single} on both real and synthetic data. 
We use the pre-trained models of Dai \etal~\cite{dai2022flare7k} and Zhang \etal~\cite{zhang2018single}.
Since Wu \etal~\cite{wu2021train} do not release their model, we retrain their network using their provided data and code.
Besides, we also compare our method with the traditional flare spot removal method~\cite{auto_removal}.
As shown in Table \ref{tab:synthetic_result}, since most of these methods are not specially designed for reflective flare removal, they cannot accurately remove the reflective flare while retaining image details simultaneously.
Although Asha \etal~\cite{auto_removal} can detect the bright spots, as shown in Fig.~\ref{fig:previous_methods}, it may also remove the light source which is quite common at night.

\noindent{\bf User Study.} To evaluate these methods in real-world situations, we also conduct a user study with 20 participants with 20 images randomly selected from our real data in Table~\ref{tab:user_study}.

\begin{table}
  \centering
  \caption{Quantitative comparison on synthetic test data.}
  \vspace{-2mm}
  \resizebox{\linewidth}{!} {
  \begin{tabular}{@{}lcccc@{}}
    \toprule
    Method & PSNR & SSIM & LPIPS& Masked PSNR \\
    \midrule
    Input  &37.30 &0.990 &0.025 &21.68\\
    Reflection removal~\cite{zhang2018single} &22.02 &0.830 &0.074 &22.88 \\
    Flare spot removal~\cite{auto_removal}   &38.29 &0.990 &0.024 &22.53 \\
    Flare removal (Wu et al.)~\cite{wu2021train} &26.13 &0.895 &0.055 &21.89 \\
    Flare removal (Dai et al.)~\cite{dai2022flare7k} &27.49 &0.942 &0.050	&21.85 \\
    Ours &\textbf{48.41}	&\textbf{0.994} &\textbf{0.004} &\textbf{32.09}\\
    \bottomrule
  \end{tabular}
  }
  \vspace{-6mm}
  \label{tab:synthetic_result}
\end{table}
\subsection{Ablation study}
\paragraph{Loss function.} 

\begin{table}
  \centering
  \caption{Ablation study for different loss functions.}
  \vspace{-2mm}
  \resizebox{0.85\linewidth}{!} {
  \begin{tabular}{@{}lcccc@{}}
    \toprule
    Loss & PSNR & SSIM & LPIPS & Masked PSNR \\
    \midrule
    w/o $\mathcal{L}_{flare}$ &47.36 &0.993 &0.004 &30.63 \\
    w/o $\mathcal{L}_{rec}$ &48.18 &0.994	&0.003 &31.33 \\
    w/o $\mathcal{L}_{vgg}$ &48.05 &0.994 &0.004 &31.39\\
    w/o $\mathcal{L}_{mask}$ &48.01 &0.993 &\textbf{0.003}	&31.03\\
    Full &\textbf{48.41}	&\textbf{0.994} &0.004 &\textbf{32.09}\\
    \bottomrule
  \end{tabular}
  }
  \vspace{-3mm}
  \label{tab:ablation_loss}
\end{table}

We conduct an ablation study to prove the necessity of each loss function used in our baseline. In our experiments, we train the model without adding $\mathcal{L}_{flare}$, $\mathcal{L}_{rec}$, $\mathcal{L}_{vgg}$ or $\mathcal{L}_{mask}$. We report the results in Table \ref{tab:ablation_loss}. The ablation study of $\mathcal{L}_{flare}$ proves the significance of bringing supervision for both flare and background images. 
Besides, when the $L_{mask}$ is removed, it degrades the reconstruction results in the flare-corrupted regions, which verifies the importance of focusing on flare regions.

\begin{figure}[t]
  \centering
   \includegraphics[width=0.9\linewidth]{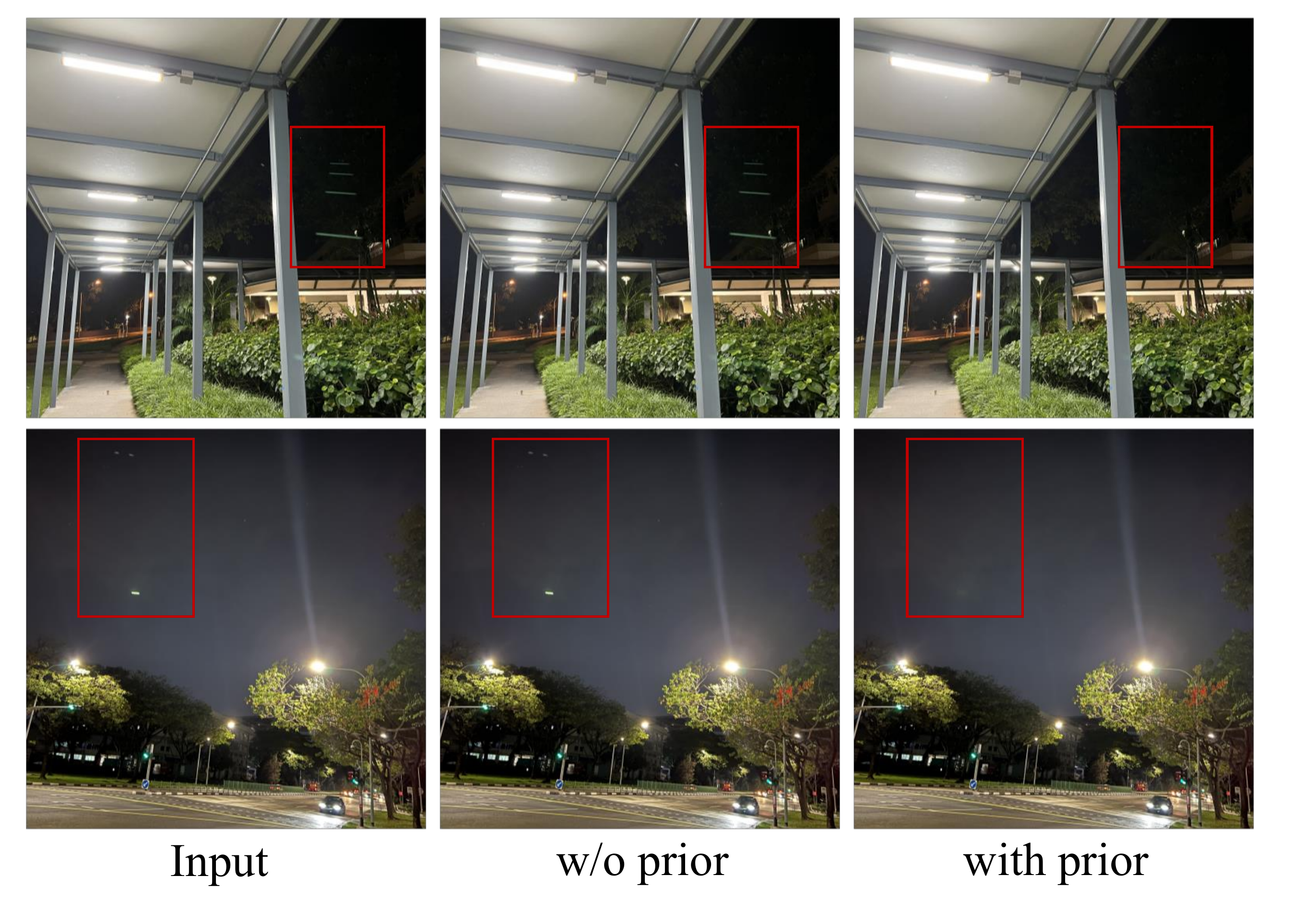}
   \vspace{-3mm}
   \caption{Visual comparison with the network's performance with and w/o optical center symmetry prior.}
   \vspace{-4mm}
   \label{fig:prior_ablation}
\end{figure}

\vspace{-4mm}
\paragraph{Prior.} 
To illustrate the performance of introducing the optical symmetry prior by concatenating a prior image, we train these baseline restoration networks with the original input. 
Since HINet~\cite{HINet} and MPRNet~\cite{MPRNet} cannot support input and output with different channels, we concatenate a black image with no information and keep other settings the same.
The quantitative results are shown in Table \ref{tab:network}, and visual comparisons on real-world images are presented in Fig. \ref{fig:prior_ablation}.
It shows that embedding this prior in the network can boost the performance of quantitative results considerably.
In Fig. \ref{fig:prior_ablation}, we observe that this prior helps the network to locate and remove the reflective flare precisely.

\begin{table}
  \centering
  \caption{Quantitative comparison for different network structures on synthetic test data. `$\dag$' indicates that the network is equipped with the proposed prior.}
  \vspace{-1mm}
  \resizebox{0.93\linewidth}{!} {
  \begin{tabular}{@{}lcccc@{}}
    \toprule
    Network & PSNR & SSIM & LPIPS & Masked PSNR \\
    \midrule
    MPRNet~\cite{MPRNet} &43.60	&0.992 &0.011 &27.64 \\
    Uformer~\cite{Uformer}  &42.57 &0.989 &0.009 &27.77 \\
    HINet~\cite{HINet} &43.78 &0.991 &0.012 &27.73 \\
    Restormer~\cite{Restormer}  &42.76 &0.991 &0.012 &27.27 \\
    \midrule
    MPRNet$^\dag$  &\textbf{48.41} &\textbf{0.994} &0.004 &\textbf{32.09} \\
    Uformer$^\dag$  &47.47 &0.991 &\textbf{0.003} &31.57 \\
    HINet$^\dag$  &48.03 &0.994	&0.003 &30.88\\
    Restormer$^\dag$ &48.11	&0.994 &0.004 &31.07 \\
    
    \bottomrule
  \end{tabular}
  }
  \label{tab:network}
  \vspace{-4mm}
\end{table}

\vspace{-3mm}
\section{Conclusion}
\vspace{-2mm}
We have presented a novel approach to tackle the challenge of removing reflective flares in nighttime photography. Our approach is inspired by the optical center symmetry prior, which suggests that reflective flares and their corresponding light sources are always symmetrical around the optical center of the lens system in smartphone cameras. Based on the observation that the patterns of reflective flares are always similar to the brightest region of the light source, we introduce a dataset named \dataset. We further propose a reflective flare removal pipeline that encodes the optical center symmetry prior into the network. Experimental results demonstrate that our proposed methods are effective in removing reflective flares in real-world scenarios and have the potential for wide-ranging applications.

\vspace{-4mm}
\paragraph{Acknowledgement:}
This study is supported under the RIE2020 Industry Alignment Fund – Industry Collaboration Projects (IAF-ICP) Funding Initiative, as well as cash and in-kind contribution from the industry partner(s).

{\small
\bibliographystyle{ieee_fullname}
\bibliography{egbib}
}

\end{document}



\title{Nighttime Smartphone Reflective Flare Removal \\Using Optical Center Symmetry Prior \\(Supplementary Material)}

\author{
Yuekun Dai$\,\,\,\,$ Yihang Luo $\,\,\,\,$ Shangchen Zhou $\,\,\, $Chongyi Li\thanks{Corresponding author.}  $\,\,\, $ Chen Change Loy \\
S-Lab, Nanyang Technological University \\
\texttt{\small \{YDAI005, c200211, s200094, chongyi.li, ccloy\}@ntu.edu.sg}\\ \vspace{-6mm}
}
\maketitle

\section{Comparison with Previous Pipelines}

To prove the performance of our dataset, we retrain Wu~\textit{et~al}.~\cite{wu2021train} and Dai~\textit{et~al}.~\cite{dai2022flare7k} on our BracketFlare dataset and conduct additional experiments as shown in Table~\ref{tab:comparison_previous}.
Wu~\textit{et~al}.~\cite{wu2021train} use U-Net as a baseline model, but we find it struggled to locate reflective flares without our prior, sometimes reducing PSNR compared to the input.
Dai~\textit{et~al}.~\cite{dai2022flare7k} set Uformer as a baseline. Since it does not encode the prior into the network, it has an obvious gap between our baseline method.
To show these methods' effects on real data, we also conduct a user study with 20 real-captured flare-corrupted images and 20 participants.
The participants are asked to identify which of these models did best at removing the reflective flares.
The experiments presented in Table.~\ref{tab:user_study2} illustrates that users strongly prefer our methods over previous methods.

\begin{table}[h]
  \centering
  \vspace{-3mm}
  \caption{Comparison of previous models retrained on our dataset.}
  \vspace{-3mm}
  \resizebox{1.0\linewidth}{!} {
  \begin{tabular}{@{}lcccc@{}}
    \toprule
    Model (Retrained) & PSNR & SSIM & LPIPS & Masked PSNR \\
    \midrule
    Input  &37.30 &0.990 &0.025 &21.68 \\
    Wu et al.~\cite{wu2021train} &31.84 &0.912 &0.032 &24.87 \\
    Dai et al.~\cite{dai2022flare7k} &42.70 &0.987 &0.010 &27.46 \\
    Ours &\textbf{48.41}	&\textbf{0.994} &\textbf{0.004} &\textbf{32.09}\\
    \bottomrule
  \end{tabular}
  }
  \vspace{-6mm}
  \label{tab:comparison_previous}
\end{table}

\begin{table}[t]
  \centering
  \caption{Percentage of users favoring our method and other retrained models.} 
  \vspace{-3mm}
  \resizebox{0.7\linewidth}{!} {
  \begin{tabular}{@{}lcccc@{}}
    \toprule
    Method & Wu et al.~\cite{wu2021train} & Dai et al.~\cite{dai2022flare7k}& Ours \\
    \midrule
    Indoor &0.5\% &3.5\%&\textbf{96\%} \\
    Outdoor &1.5\% &2.5\%&\textbf{96\%} \\
    \bottomrule
  \end{tabular}
  }
  \vspace{-4mm}
  \label{tab:user_study2}
\end{table}

\section{Flare Removal for Downstream Tasks}

\begin{figure}[t]
  \centering
  \includegraphics[width=0.49\textwidth]{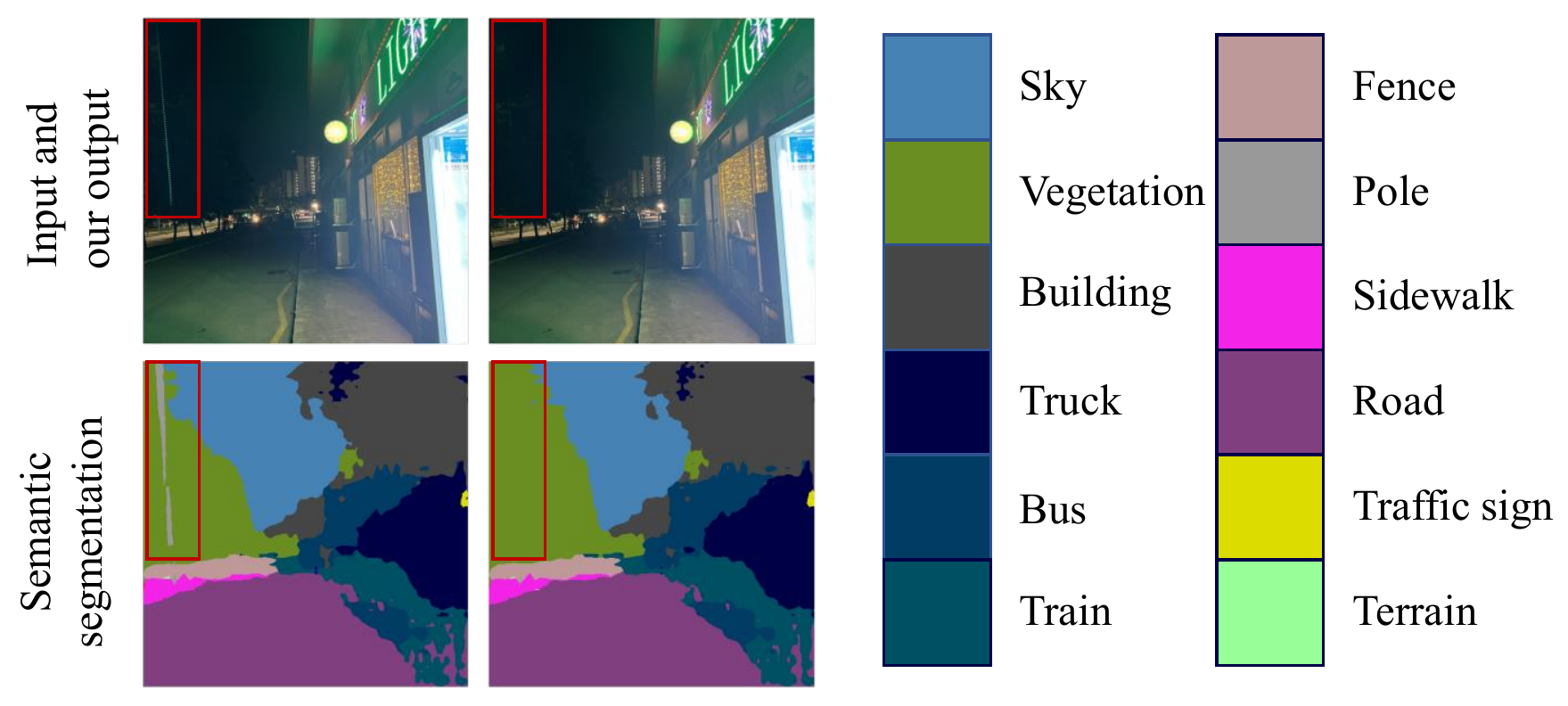}
   \vspace{-6mm}
   \caption{Visual comparison of estimated segmentation for real-world flare-corrupted and flare-removed image pairs. The segmentation maps are calculated by DANNet~\cite{wu2021dannet}, a nighttime semantic segmentation algorithm. }
   \vspace{-7mm}
   \label{fig:seg}
\end{figure}

Fig.~\ref{fig:seg} demonstrates that it can benefit downstream tasks such as image segmentation. 
In the figure, it can be observed that the reflective flares are misclassified as a pole, which may pose potential risks for nighttime driving. 
Removing such reflective flares can help achieve more robust and reliable results, and ultimately benefit the users.

\begin{figure*}[h]
  \centering
   \includegraphics[width=0.9\linewidth]{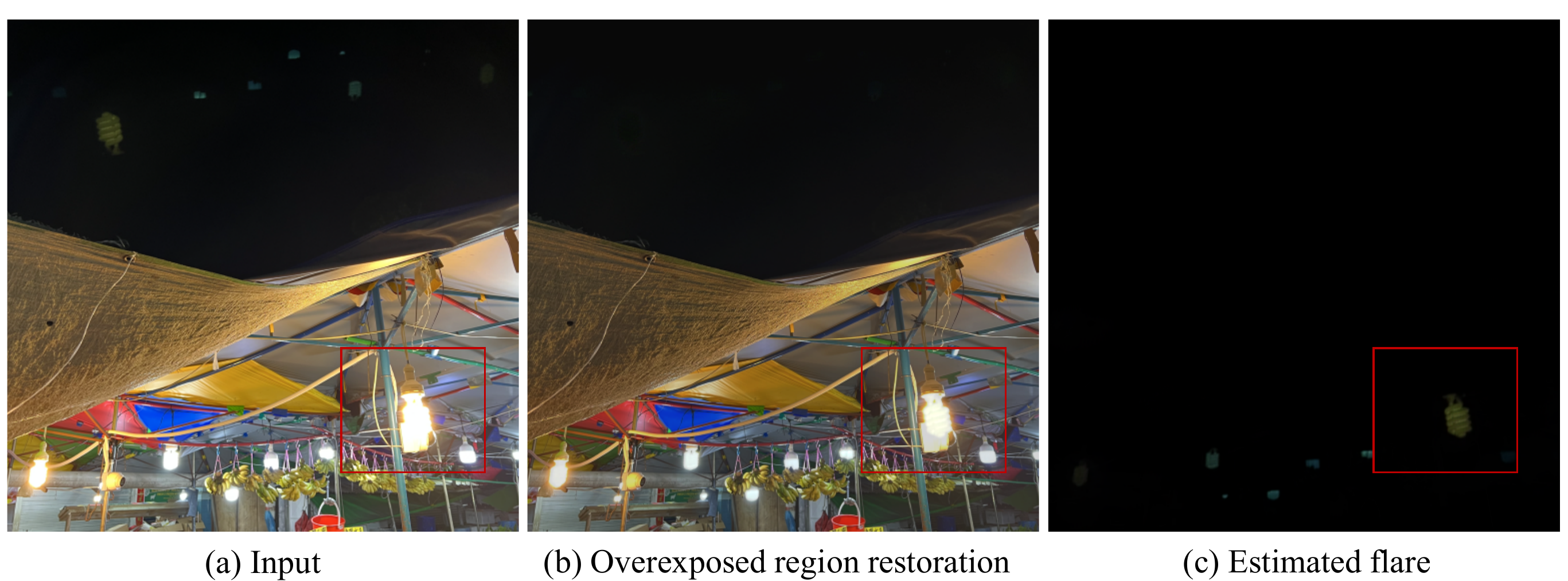}
   \caption{Our reflective flare removal method can help achieve overexposed region restoration. As shown in  (c), the details of the light source can be reconstructed well by referring to the estimated flare.}
   \vspace{-3mm}
   \label{fig:HDR}
\end{figure*}

\section{Potential in Overexposure Restoration}
Due to the limited dynamic range of some cameras, image details especially in light source regions are always saturated and difficult to recover.
To achieve the details of the overexposed regions, the mainstream solution is to use HDR (high-dynamic range) imaging with multi-exposure capture~\cite{debevec2008recovering, reinhard2010high}.
%
%
Our main idea may provide a potential solution to this issue.

Specifically, since smartphone reflective flare can be considered as a short-exposure image, our method also provides the potential in implementing overexposed regions restoration.
As shown in Fig.~\ref{fig:HDR}, we can separate a reflective flare and rotate the estimated flare 180 degrees around the optical center to match the flare with the light source.
We suppose the exposure step between the flare and the original input is 12 EV.
Then, these two images with low-dynamic ranges will be merged to generate an HDR image with clear details in the light source.
The visualization results in Fig.~\ref{fig:HDR} show that our method can recover the details of the saturated regions well.

\section{More Visual Results}
\begin{figure*}[t]
  \centering

   \includegraphics[width=0.9\linewidth]{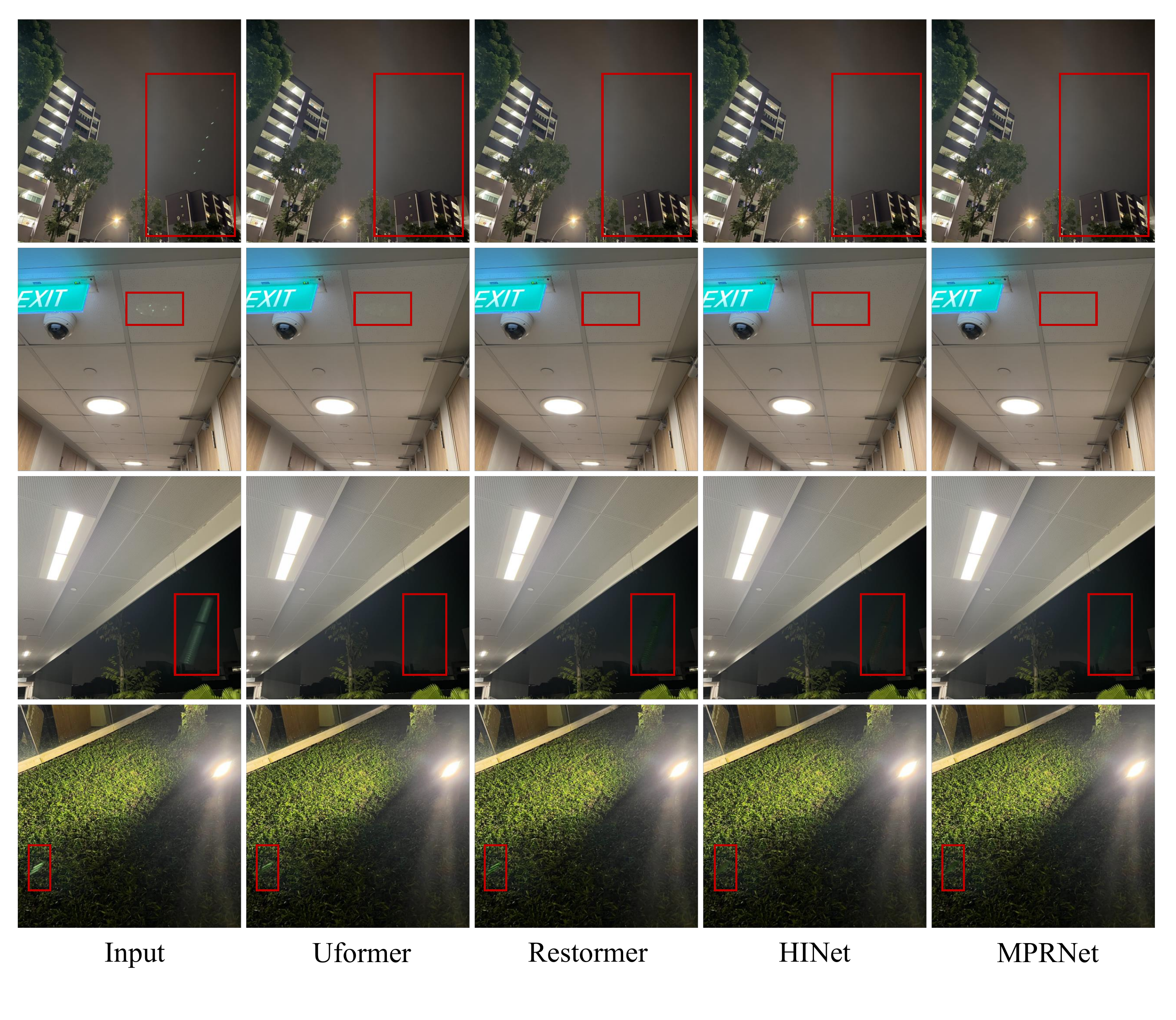}
    \vspace{-8mm}
   \caption{Visual comparison of different networks retrained on our dataset for restoring the real-world flare-corrupted images. These image restoration networks include Uformer~\cite{Uformer}, Restormer~\cite{Restormer}, HINet~\cite{HINet}, and MPRNet~\cite{MPRNet}.}
   \label{fig:different_nets}
\end{figure*}

To further demonstrate the effectiveness of our dataset, 
we retrain different neural networks including Uformer~\cite{Uformer}, Restormer~\cite{Restormer}, HINet~\cite{HINet}, and MPRNet~\cite{MPRNet} using our proposed dataset. 
The visual results of different networks for restoring real-world images are presented in Fig.~\ref{fig:different_nets}. 
From the visual comparison, we can observe that all the networks retrained on our dataset can remove the flares. Among them, 
MPRNet~\cite{MPRNet}  obtains the best performance on flare-corrupted regions, as highlighted in the red boxes of  Fig.~\ref{fig:different_nets}.
The results manifest the effectiveness of our proposed dataset.

\begin{figure*}[t]
  \centering

   \includegraphics[width=0.90\linewidth]{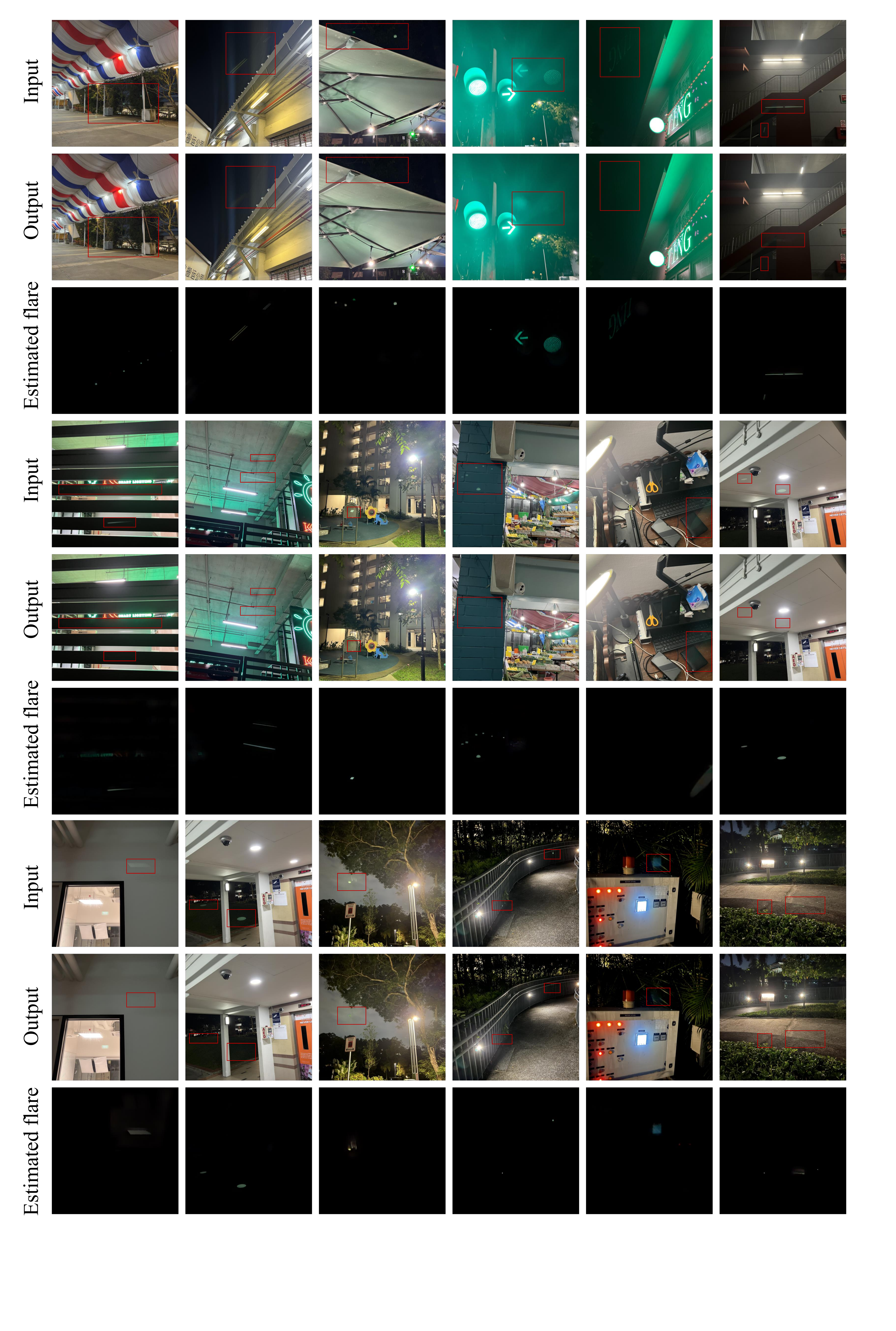}
   \vspace{-20mm}
   \caption{Our results on real-world nighttime flare-corrupted images. In this figure, we use MPRNet~\cite{MPRNet} as  the baseline method to estimate the flare and output image. }
   \label{fig:more_results1}
\end{figure*}

\begin{figure*}[t]
  \centering

   \includegraphics[width=0.90\linewidth]{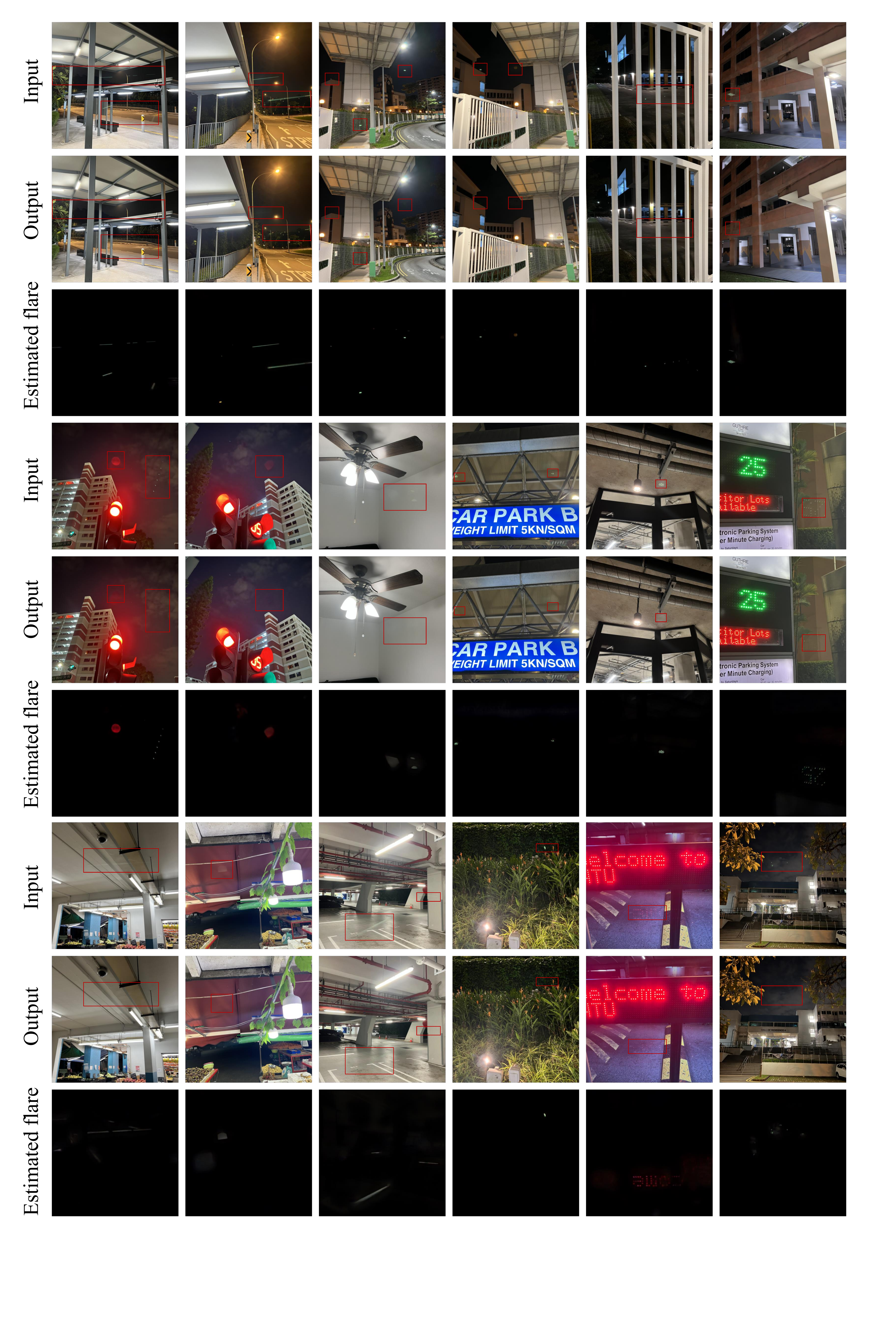}
   \vspace{-20mm}
   \caption{More results on real-world nighttime flare-corrupted images. In this figure, We use MPRNet~\cite{MPRNet} as the baseline method to estimate the flare and output image.}
   \label{fig:more_results2}
\end{figure*}

To show our dataset and prior's generalization ability in different real-world scenes, we show more results of the flare-corrupted images captured by iPhone 13 Pro that is known for severe reflective flares in Fig.~\ref{fig:more_results1} and Fig.~\ref{fig:more_results2}.
The results show that our proposed method can tackle a huge diversity of light sources and achieve good visual performance in different scenes.
It is owing to the versatility of our proposed prior, as well as the diversity and balanced distribution of our proposed dataset.

We also capture many flare-corrupted video clips by Huawei P40, iPhone 13 Pro, and ZTE Axon 20 5G.
To prevent triggering the anti-shake module of the smartphone, we try to keep the smartphone from large movement.
Then, we process these videos with our method frame by frame.
The video results can be found in a separate video demo.
The video demo shows that our method can achieve robust and consistent flare-removal results even for real-world videos.

\section{Limitation}
As shown in Fig. 10 of the main paper, our method can generalize well to different types of smartphones. 
However, it is based on the fact that most smartphone cameras satisfy the optical center symmetry prior.
For some professional cameras, this prior does not always hold.
Besides, due to the dispersion of thick lenses in professional cameras, the main flare spot of the professional cameras are not always the same as the light source's pattern.
The solutions for these cases will be left for future work.

{\small
\bibliographystyle{ieee_fullname}
\bibliography{egbib}
}